\documentclass[runningheads, orivec]{llncs}

\usepackage[utf8]{inputenc} 
\usepackage[T1]{fontenc}    
\usepackage{hyperref}       
\usepackage{url}            
\usepackage{booktabs}       
\usepackage{amsfonts}       
\usepackage{nicefrac}       
\usepackage{microtype}      
\usepackage{lipsum}		
\usepackage{graphicx}
\usepackage{doi}

\makeatletter
\edef\orig@output{\the\output}
\output{\setbox\@cclv\vbox{\unvbox\@cclv\vspace{0pt plus 20pt}}\orig@output}
\makeatother

\usepackage{amsmath,amssymb,amsfonts,amsbsy}
\usepackage{algpseudocode}
\usepackage{algorithm}
\usepackage{graphicx}
\usepackage{listings}
\usepackage{textcomp}
\usepackage{xcolor}
\usepackage{soul}
\usepackage{float}
\usepackage{comment}
\usepackage{subfigure}
\usepackage{cite}

\usepackage{tikz}
\usetikzlibrary{shapes,arrows,shadows,positioning}
\usetikzlibrary{matrix,chains,positioning,decorations.pathreplacing}

\newcommand{\bs}[1]{\boldsymbol{#1}}

\newcommand{\hlb}[1]{\textcolor{blue}{#1}}

\title{Effect sizes as a statistical feature-selector-based learning to detect breast cancer}

\author{Nicolás Masino\inst{1} \and Antonio~Quintero-Rinc\'on\inst{1,2}}
\institute{Data Science and AI Laboratory\inst{1}, Data Science Department\inst{1}, \\Computer Science Department\inst{2} \\ Catholic University of Argentina (UCA), Argentina \\
\email{nicolasmasino@uca.edu.ar, antonioquintero@uca.edu.ar}\\
To cite this work, please use this reference
\doi{10.1109/ARGENCON62399.2024.10735908}}

\begin{document}
\maketitle
\begin{abstract}
Breast cancer detection is still an open research field, despite a tremendous effort devoted to work in this area. Effect size is a statistical concept that measures the strength of the relationship between two variables on a numeric scale. Feature selection is widely used to reduce the dimensionality of data by selecting only a subset of predictor variables to improve a learning model. In this work, an algorithm and experimental results demonstrate the feasibility of developing a statistical feature-selector-based learning tool capable of reducing the data dimensionality using parametric effect size measures from features extracted from cell nuclei images. The SVM classifier with a linear kernel as a learning tool achieved an accuracy of over 90\%. These excellent results suggest that the effect size is within the standards of the feature-selector methods.

\end{abstract}
%

\section{Introduction}
Breast cancer is a disease frequently diagnosed in women in which abnormal breast cells grow uncontrollably until tumors form. According to the World Health Organization \cite{WHO2024} in 2022, breast cancer was diagnosed to over 2.3 million women and 670,000 deaths globally. Cancer treatment is patient-specific through radiation therapy, medications, and surgery; and depends on the type of cancer and its spread in the body. 

Effect size (EF) is an association-magnitude measure between two populations under research. EF has become more popular in recent years in meta-analyses of psychological, educational, and behavioral treatments \cite{NewStatistics2024}. EF can be estimated through parametric or non-parametric kernels. EF expresses across a numerical decision rule the practical significance or strength of a research outcome. This numerical scale yields an index that lets us know how meaningful the relationship between variables or the difference between groups is. It is common knowledge that statistical significance denoted by p-values affects an outcome, while effect sizes represent practical significance. A large effect size means that the research outcome has practical importance, while a small effect size indicates limited practical applications. The family of EF indices can be separated into two types of measures. The standardized differences between two groups and the correlation measure of effect size \cite{Cohen1998}. This work is focused on estimating the parametric standardized difference between two independent populations vs a dichotomous dependent variable.

The well-known free Diagnostic Wisconsin Breast Cancer database hosted in the Machine Learning Repository at the University of California, Irvine \cite{dataset} was considered for experimentation purposes. Its features describe the characteristics of the cell nuclei present in a digitized image of a fine needle aspirate (FNA) of a breast mass. For details, consult the Section \ref{ssec:data} introduced below. 

This work proposes a statistical feature-selector-based learning to detect breast cancer through effect size.
The underlying idea of feature-selector-based learning lies in finding a good interaction between the features and selecting only the most relevant. In this context, a correct combination of features may provide higher predictive power and increase the precision of a learning model with minimal risk. Therefore, performing feature-selector-based learning allows for identifying an optimal feature combination and a dimensional data reduction to improve the predictive learning capacity. This process has several advantages. It can significantly reduce model training time and prediction speed during production deployment. Additionally, it makes the model less complex and easier to explain. 

Feature-selection methods can be grouped into three approaches: Filter method, Wrapper method, and Embedded method \cite{li2017, jundong2017}. Filter methods are based on a relevance index based on correlation coefficients or test statistics. The methods used are Correlation-based filters, Relevance indices based on Distances between distributions, or Information Theory, Decision trees for ranking, or Reliability and Bias of relevance indices. Wrapper's methods utilize the performance of a learning machine trained using a given feature subset. The methods used are based on forward selection and backward elimination. Both methods use search strategies to explore the space of all possible feature combinations. Embedded methods incorporate feature subset generation and evaluation in the training algorithm \cite{jundong2017, miao2016}. The methods used are based on forward selection and backward elimination with an optimization of scaling factors. Embedded and wrapper methods perform well for a given classifier. However, these approaches represent more computational complexity, above all the data dimension is high \cite{saraswat2014, Remeseiro2019, li2017}. Note that, the proposed EF as statistical feature-selection-based learning fits the filter method. 
Many well-known feature selection techniques are within the filter, wrapper, or embedded methods. That is why choosing a specific technique is not easy. It is essential to know the data dimension, the size of the data, and the computational cost that is required and accepted.
In medical diagnosis scenarios, the most common techniques are Correlation-Based Feature Selection, Consistency-Based Filter, INTERACT, Information Gain, Relief, Recursive Feature Elimination, and Lasso Regularization \cite{Remeseiro2019}. Specifically in feature selection to identify relevant cell nuclei features, a variety of studies can be found to diagnose breast cancer, for example, in \cite{Cheng2018} three different techniques were proposed: minimum redundancy maximum relevance, Wilcoxon's rank-sum test, and Random Forest, in \cite{Osareh2010} by employing a Sequential Forward Selection (SBS), or across an embedded approach using SVM based on the F-score to predict it \cite{akay2009}. Another study from the same field compared Random Forest (RF) with other select feature procedures, like SVM-RF, RRF, SBS, and VarSelRF. Binary Random Forest Feature Selection method (BRFFS) was proposed to reduce the nuclear features and classify leukocytes \cite{saraswat2014}. For a comprehensive treatment of feature selection see \cite{Guyon2006,Haque2023}.

The hypothesis to be assessed in this work is under the assumption that the behavior of a variable is different for each class. Specifically in breast cancer scenarios, the features should have different values for their classes, called malignant cancerous and benign non-cancerous samples. To assess this hypothesis, the parametric measure of effect size was proposed to quantify the difference in the distributions. Note that the techniques used in this study are all well-known in the scientific community. Still, to our knowledge, the effect size measurement has never been used as a statistical feature-selector-based learning scheme. This is the main contribution of this work.

The remainder of this paper is organized as follows. Section
\ref{sec:met} presents the proposed method and provides an introduction of the dataset used in Section \ref{ssec:data}, the effect size in Section \ref{ssec:effect}, the feature-selector-based decision rule in Section \ref{ssec:rule}, the feature-selector-based learning with a support vector machine in Section \ref{ssec:learning}, and the confidence intervals in section \ref{ssec:ci}. In Section \ref{sec:res}, the results are discussed and analyzed. Finally in Section \ref{sec:con} conclusion, advantages, limitations, and future works are given.

\section{Methodology}
\label{sec:met}
The pipeline proposal is as follows. Initial data with the features of cell nuclei present from digitized images obtained by FNA of a breast mass were given. Then, the parametric effect size measurement is computed for each sample to estimate a dimensional feature reduction from the original feature data. Feature-selector-based learning is calculated according to the values of a numerical decision rule. If the observed value of the effect size is greater than a numerical value, then it is decided if the feature is significant or not. The numerical value is chosen for the standardized effect size scale value, or estimating the mean of the rank of their values. Finally, with the new data arising from feature reduction, a classification-based learning scheme for detecting breast cancer is computed in such a way that can be utilized in prevalent clinical settings. The dataset, software, parametric effect size methods, feature selector-based decision rule, and feature selector-based learning are then presented.

\subsection{Dataset}
\label{ssec:data}
For experimentation purposes, the Diagnostic Wisconsin Breast Cancer Database from the University of California, Irvin, was considered \cite{Street1993, dataset}. This dataset is based on image-processing techniques that use custom active contour models, known as snakes. Images close to the boundaries of a set of cell nuclei are selected using a fine needle aspiration slide. Thus, size, shape, and texture can be acquired precisely because the snakes are deformed to the exact shape of the nuclei. For each nucleus, $10$ features called: radius, perimeter, area, texture, compactness, smoothness, concavity, concave points, symmetry, and fractal dimension, were estimated. Also, the mean value, the largest or worst value, and the standard error of each feature are found over the range of isolated cells. Different combinations of features were then tested to find those that best discriminated between benign and malignant samples, yielding $30$ features in total related to mean texture, worst area, and worst smoothness, see Table~\ref{tab:effectsizes}. The dataset contains $569$ binary observations that were split into two groups. $212$ malignant cancerous samples, and $357$ benign non-cancerous samples. See \cite{Street1993} for a detailed description, and explanation of this dataset.

\subsection{Software}
\label{ssec:soft}
The implementations were implemented through RStudio 2023.09.1+494 "Desert Sunflower" Release, using the Learning Statistics with R (LSR) library and MBESS Package.

\subsection{Effect size}
\label{ssec:effect}
Let $\boldsymbol{M} \in \mathbb{R}^{F\times N}$ denote the matrix gathering with $F$ raw features and $N$ values extracted from cell nuclei images with two groups related to the malignant cancerous and benign non-cancerous samples. Let $M_i \in \mathbb{R}^{1\times N_{1,i}}$ the malignant cancerous sample vector, and $B_i \in \mathbb{R}^{1\times N_{2,i}}$ the benign non-cancerous sample vector for each $i$ feature, with $1\leq i \leq F$, $N_1\leq N$, and $N_2\leq N$ respectively. For a detailed explanation of the estimation of the effect size and its interpretation, we refer the reader to \cite{Cohen1998, NewStatistics2012, NewStatistics2024}

\subsubsection{Cohen's d standardized effect size}
\label{ssec:ef}
Cohen's d standardized effect size \cite{Cohen1998} is defined as:
\begin{align}
    \label{eq:cohenEq}
    d &= \frac{\overline{M}_i-\overline{B}_i}{\sqrt{\frac{(N_{1,i}-1)SD^2_{N_{1,i}}+(N_{2,i}-1)SD^2_{N_{2,i}}}{N_{1,i}+N_{2,i}-2}}}
\end{align}
where $\overline{\bullet}$ is the mean for each group, $SD$ is the pooled standard deviation, $N_{1,i}$ and $N_{2,i}$ are the samples size for each $i$ feature. Note that the standard deviation is estimated from the differences between each observation and the mean for the group. These differences are the sum of squares used to avoid the positive and negative values from canceling each other out and summing. This value is divided by the number of observations minus one, which is Bessel's correction for bias in the population calculation variance based on the least squares estimate \cite{Lakens2013}. Finally, the square root is computed. 
Since Cohen's d expressed the effect size for t-test results in units of variability, two assumptions must be considered: the first is related to the normal assumption and the second is the assumption of homogeneity of variance.

\subsubsection{Cohen's D}
\label{ssec:cohenDD}
Cohen's D is the Cohen's d without homogeneity of variance assumption, also known as Welch's t-test. It can be formulated as:
\begin{align}
    D = \frac{\overline{M_i}-\overline{B_i}}{\sqrt{\frac{SD^2_{N_{1,i}} + SD^2_{N_{2,i}}}{2}}}
    \label{eq:cohenDD}
\end{align}

\subsubsection{$U$ measures}
\label{ssec:U measures}
Let $\Phi$ be the standard normal cumulative distribution function. Cohen's $U_3$ represents the percentage of $M_i$ is upper half of the cases of the $B_i$, is computed as:
\begin{equation}
    U_3 = \Phi(d)
    \label{eq:cohenU3}
\end{equation}

In the same way, Cohen's $U_2$ is defined as:
\begin{equation}
    U_2 = \Phi\left(\frac{d}{2}\right)
    \label{eq:cohenU2}
\end{equation}
Cohen's $U_2$ measures the percentage in $B_i$ that exceeds the same percentage in $M_i$.
Finally, Cohen's $U_1$ is the following non-overlapping percentage:
\begin{equation}
    U_1 = \frac{2U_2-1}{U_2}
    \label{eq:cohenU1}
\end{equation}
Cohen's $U_1$ is the amount of combined area not shared by the two population distributions, $M_i$ and $B_i$ \cite{Cohen1998}.

$U$ measures lie between the values $0$ and $1$. A value $0$ means a full overlapping or null effect, while a value $1$ means a large effect. Hence the variables whose $U$ measures are closer to $1$ are better at classifying among benign and malignant samples.

\subsection{Feature-selector-based decision rule}
\label{ssec:rule}
The effect size value (or strength) between two samples, such as the cancerous or benign non-cancerous samples, can be interpreted through the numerical decision rule presented in Table~\ref{tab:scale}. Precisely, the observed value from this measure is the statistical value calculated that allows a reduction in the dimensionality of data.

\begin{table}[htbp]
\caption{Decision rules for assessing the strength of the effect size measurement of the observed difference. $\mu$ is the mean.}
\begin{center}
\begin{tabular}{|c|c|}
\hline
\textbf{Observed value for Cohen's d and Cohen's D} & \textbf{Strength} \\
\hline
 Effect size value $< 0.2$ & Null \\
\hline
$0.2 \leq$ Effect size value $< 0.5$ & Small\\
\hline
$0.5 \leq$ Effect size value $\leq 0.8$ & Medium\\
\hline  
Effect size value $>~0.8$ & Large\\
\hline
\textbf{Observed value for Cohen's $\bs{U_1}$ to Cohen's $\bs{U_3}$} & \textbf{Strength} \\
\hline
 Effect size value $<\mu$(rank(Effect size values)) & Null \\
\hline
Effect size value $\geq\mu$(rank(Effect size values))& Large\\
\hline
\end{tabular}
\label{tab:scale}
\end{center}
\end{table}

\subsection{Feature-selector-based learning}
\label{ssec:learning}
\begin{figure}[htbp]
    \centering
    \includegraphics[width=0.66\linewidth]{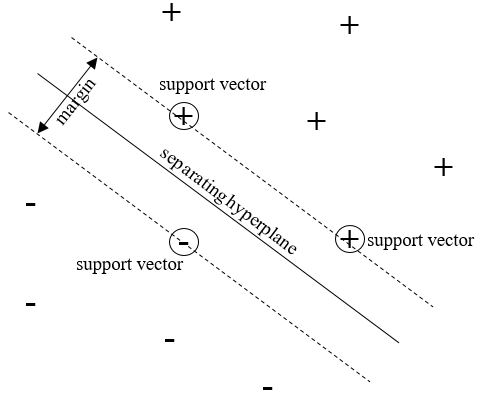}
    \caption{Linear-SVM illustration}
    \label{fig:svm}
\end{figure}

Support Vector Machines (SVM) are based on finding a hyperplane that separates the classes under study at the same distance. SVM results in an optimization problem whose primary purpose is to find the largest number of margins between the points in space and the separating hyperplane. The points within these margins are called support vectors. Fig.~\ref{fig:svm} illustrates how a Support Vector Machine works.

 Let a set with $N$ observations value in bi-dimensional space, with order pairs $(x_1, x_2)$. Let a variable target with two-factor levels $Y_i \in \{+1,-1\}$. Then the separating hyperplane is defined as:
\begin{align}
    \overline{w}.\overline{x_i}+b &=0\\
     x_i &= \overline{x_\perp} + r\frac{\overline{w}}{||\overline{w}||}
\label{eq:margin}
\end{align}
The expression \eqref{eq:margin} is the distance of a point to the decision boundary, where $r$ is the distance of $x_i$ from the decision boundary whose normal vector is $\overline{w}$, and $\overline{x_\perp}$ is the orthogonal projection of $x_i$ onto this boundary. See \cite{Murphy2022, Tibshirani2023} for a comprehensive treatment of the properties of the SVM.

\subsection{Confidence intervals (CIs)}
\label{ssec:ci}
CIs are estimated using the non-centrality parameter (NCP) method. This is a pivot method that finds the NCP of a non-central $t$, $F$, or $\chi^2$ distribution that places the observed $t$, $F$, or $\chi^2$ test statistic at the desired probability point of the distribution \cite{Kelley2008}. Furthermore, the NCP method is useful to find the confidence intervals for Cohen's d and Cohen's D, but not for the $U$ measures. For these measures, the Bootstrap method is used \cite{boot, Tibshirani2023}. 
After estimating these confidence bounds on the NCP, they are converted into the effect size metric to yield a confidence interval for the effect size.

Some equations are necessary to express the relationship between effect size and NCP. Please note that equation \eqref{eq:t2d} is given the Cohen expression \eqref{eq:cohenEq} and the t-test expression defined in \eqref{eq:ncpEstimated}, therefore it is possible to convert the t-test statistic into a Cohen's d value.
\begin{align}
    \label{eq:t2d}
    d = t\sqrt{\frac{N_{1,i}+N_{2,i}}{N_{1,i}N_{2,i}}}
\end{align}

In such case, especially for the standardized mean difference, the population NCP for the two independent groups $t$-test is defined as:
\begin{align}
    \label{eq:ncp2groups}
    \lambda = \frac{\mu_{1,i}-\mu_{2,i}}{\sigma_i\sqrt{\frac{1}{N_{1,i}}+\frac{1}{N_{2,i}}}}
\end{align}

Where $\mu_{1,i}$ and $\mu_{2,i}$ are the population means for the malignant and benign samples for each $i$ feature. Since Cohen's d assumes homogeneous variances, both $\sigma^2_{1,i}$ and $\sigma^2_{2,i}$ are identical, so the populations mean the difference is divided by the product between $\sigma_i$ and the square root.
The population NCP is estimated as follows:
\begin{align}
\label{eq:ncpEstimated}
    \hat{\lambda} = \frac{\overline{M_i}-\overline{B_i}}{SD_i\sqrt{\frac{1}{N_{1,i}}+\frac{1}{N_{2,i}}}}
\end{align}

Note that $\hat{\lambda}$ equals the observed $t$-test statistic. For the estimated non-centrality parameter, $SD$ is the pooled deviation defined in \eqref{eq:cohenEq}. 
Given the NCP equations, it is possible to compute the confidence intervals for $\lambda$. Let $p$, the probability that the feature is contained within a random interval is $1-\alpha$, where $\alpha$ is the Type I error rate and $1-\alpha$ is the confidence level coverage, and let $T$ be the standardized of the effect size. The confidence intervals are estimated using the following equation, with $\nu$ degrees of freedom.

\begin{align}
    \label{eq:CI}
    p\left[
    t_{(\alpha/2;\nu)} \leq T \leq t_{(1-\alpha/2;\nu)}
    \right] = 1-\alpha
\end{align}

For Cohen's d, $T$ is given as:
\begin{align}
    \label{eq:cohendT}
    T =  \frac{(\overline{M_i}-\overline{B_i})-(\mu_{1,i}-\mu_{2,i})}{SD_i\sqrt{\frac{1}{N_{1,i}}+\frac{1}{N_{2,i}}}}
\end{align}

For Cohen's D, $T$ is given as:
\begin{align}
    \label{eq:cohenDT}
    T = \frac{(\overline{M_i}-\overline{B_i})-(\mu_{1,i}-\mu_{2,i})}{\sqrt{\frac{SD^2_{1,i}}{N_{1,i}}+\frac{SD^2_{2,i}}{N_{2,i}}}}
\end{align}

To get the confidence intervals for Cohen's d, it is necessary to find the confidence intervals for $\lambda$, and then the bounds are transformed to the Cohen's d scale using \eqref{eq:t2d}. The CIs can therefore be written as:
\begin{align}
\label{eq:lambda1}
    p[\lambda_L \leq \lambda \leq \lambda_U]=1-\alpha
\end{align}

The value $\lambda_L$ is found such that $p(\hat{\lambda}|\lambda_L)=\alpha/2$ and $\lambda_U$ is found such that $p(\hat{\lambda}|\lambda_U)=1-\alpha/2$ with $\nu=N_{1,i}+N_{2,i}-2$ degrees of freedom. Thus, the confidence intervals are finally defined as:
\begin{align}
\label{eq:lambda2}
    \left[\lambda_L\sqrt{\frac{N_{1,i}+N_{2,i}}{N_{1,i}N_{2,i}}}; \lambda_U\sqrt{\frac{N_{1,i}+N_{2,i}}{N_{1,i}N_{2,i}}}\right]
\end{align}

The Cohen's D procedure is similar, but using $T$ based on Welch's $t$-test. Note that its standard deviation is a mean of the standard deviations of the benign and malignant groups, so the equation~\eqref{eq:t2d} should be written as:
\begin{align}
\label{eq:cohendf}
    d=\frac{t\sqrt{\frac{SD^2_{1,i}}{N_{1,i}}+\frac{SD^2_{2,i}}{N_{2,i}}}}{\sqrt{\frac{SD^2_{1,i}+SD^2_{2,i}}{2}}}
\end{align}
 
Moreover, the degrees of freedom are approximated by using the Welch-Satterthwaite equation \cite{ahad2014}:
\begin{align}
\label{eq:ws}
    \nu = \frac{\left(\frac{SD^2_{1,i}}{N_{1,i}}+\frac{SD^2_{2,i}}{N_{2,i}}\right)^2}{\frac{\left(\frac{SD^2_{1,i}}{N_{1,i}}\right)^2}{N_{1,i}-1}+\frac{\left(\frac{SD^2_{2,i}}{N_{2,i}}\right)^2}{N_{2,i}-1}}
\end{align}

Confidence intervals for the $U$ measures are estimated using the Bootstrap method. The main idea is to take $B$ random samples from the original sample, just as samples are taken from the population. This allows us to create a Bootstrap distribution for $\hat{\theta}^*$ or a Bootstrap estimate that follows the sampling distribution for $\hat{\theta}$. Therefore, the goal is to estimate the distribution of $\frac{\hat{\theta}-\theta}{SD(\hat{\theta})}$ so that confidence intervals can be constructed, for this the equation \eqref{eq:CI} is used, but defining $T$ as:
\begin{align}
\label{eq:bootT}
    T^*_b = \frac{\hat{\theta}^*_b-\hat{\theta}}{SD(\hat{\theta}^*_b)}
\end{align}
where $\hat{\theta}^*_b$ is the value of $\hat{\theta}$ for the $b^{th}$ bootstrap sample. Likewise, $\hat{\theta}$ is the effect size of interest. The below algorithm represents the Bootstrap procedure.

\begin{algorithm}
\caption{Confidence intervals for U measures.}
\label{alg:Umesures}
\begin{algorithmic}
\Require  $\mathbf{X} = \{x_1, x_2, \ldots, x_n\}$ \Comment{Data}
\Require $B_1$ \Comment{Bootstrap samples numbers for $\hat{\theta}^*$}
\Require $B_2$ \Comment{Bootstrap samples numbers for $SD(\hat{\theta}^*)$}
\Require $1 - \alpha$ \Comment{Confidence level}
\For{$b = 1$ to $B_1$}
    \State $\mathbf{X}^*_b = \{x^*_1, x^*_2, \ldots, x^*_n\}$ \Comment{Generate $B_1$ Bootstrap sample from $\mathbf{X}$}
    \State Compute $\hat{\theta}^*_b$
\EndFor
\For{$b = 1$ to $B_2$}
    \State $\mathbf{X}^*_b = \{x^*_1, x^*_2, \ldots, x^*_n\}$
    \Comment{Generate $B_2$ Bootstrap sample from $\mathbf{X}$}
    \State Compute $SD(\hat{\theta}^*_b)$
\EndFor
\State Sort Bootstrap values $\{\hat{\theta}^*_1, \hat{\theta}^*_2, \ldots, \hat{\theta}^*_B\}$ in ascending order 
\State Compute $T^*_b$ and find the percentiles $\hat{t}^*_{\alpha/2}$ and $\hat{t}^*_{1-\alpha/2}$

\Return $[\hat{\theta}-\hat{t}^*_{1-\alpha/2}SD(\hat{\theta}^*); \hat{\theta}-\hat{t}^*_{\alpha/2}SD(\hat{\theta}^*)]$
\end{algorithmic}
\end{algorithm}

\section{Results and Discussion}
\label{sec:res}

\begin{figure}[htbp]
\centering
\subfigure{\includegraphics[width=82mm]{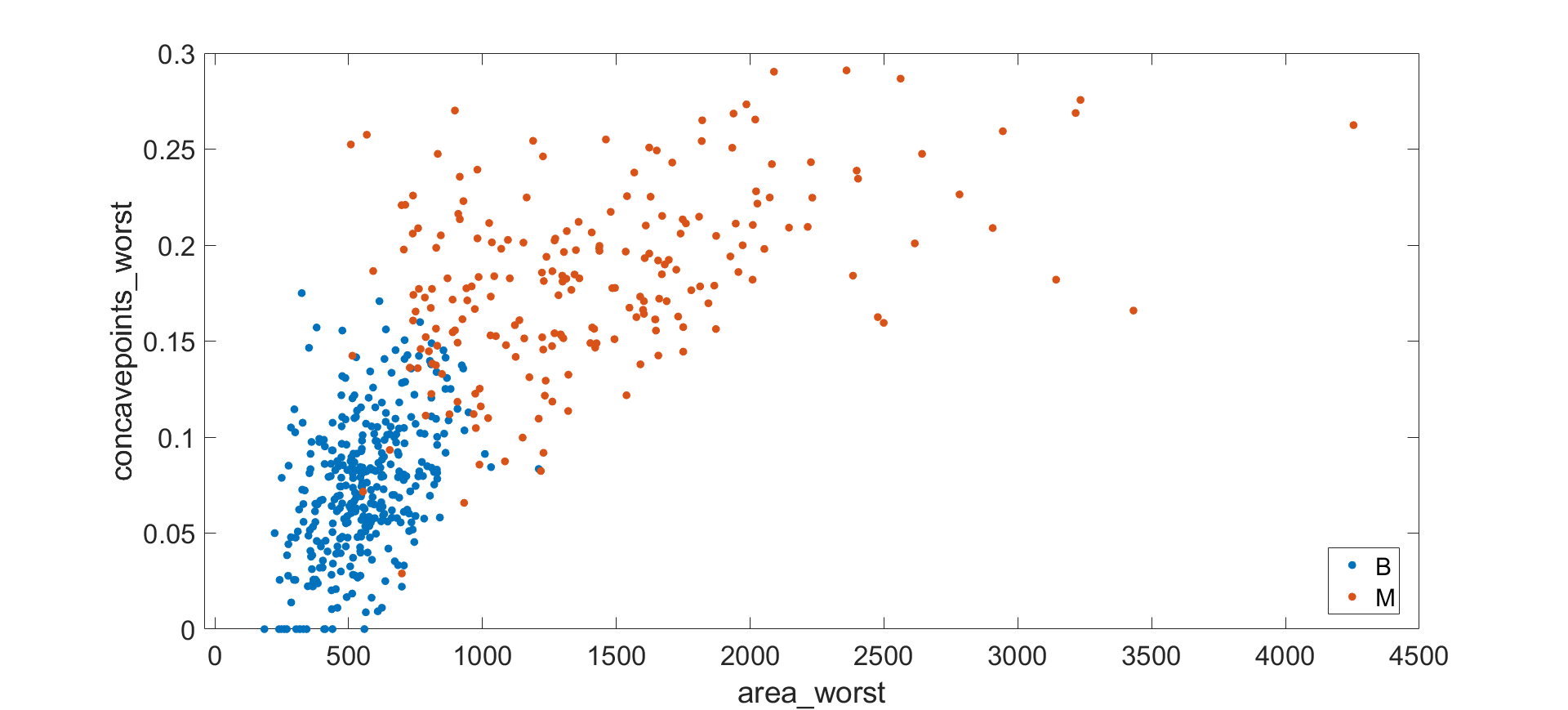}\label{fig:w1}}
\subfigure{\includegraphics[width=82mm]{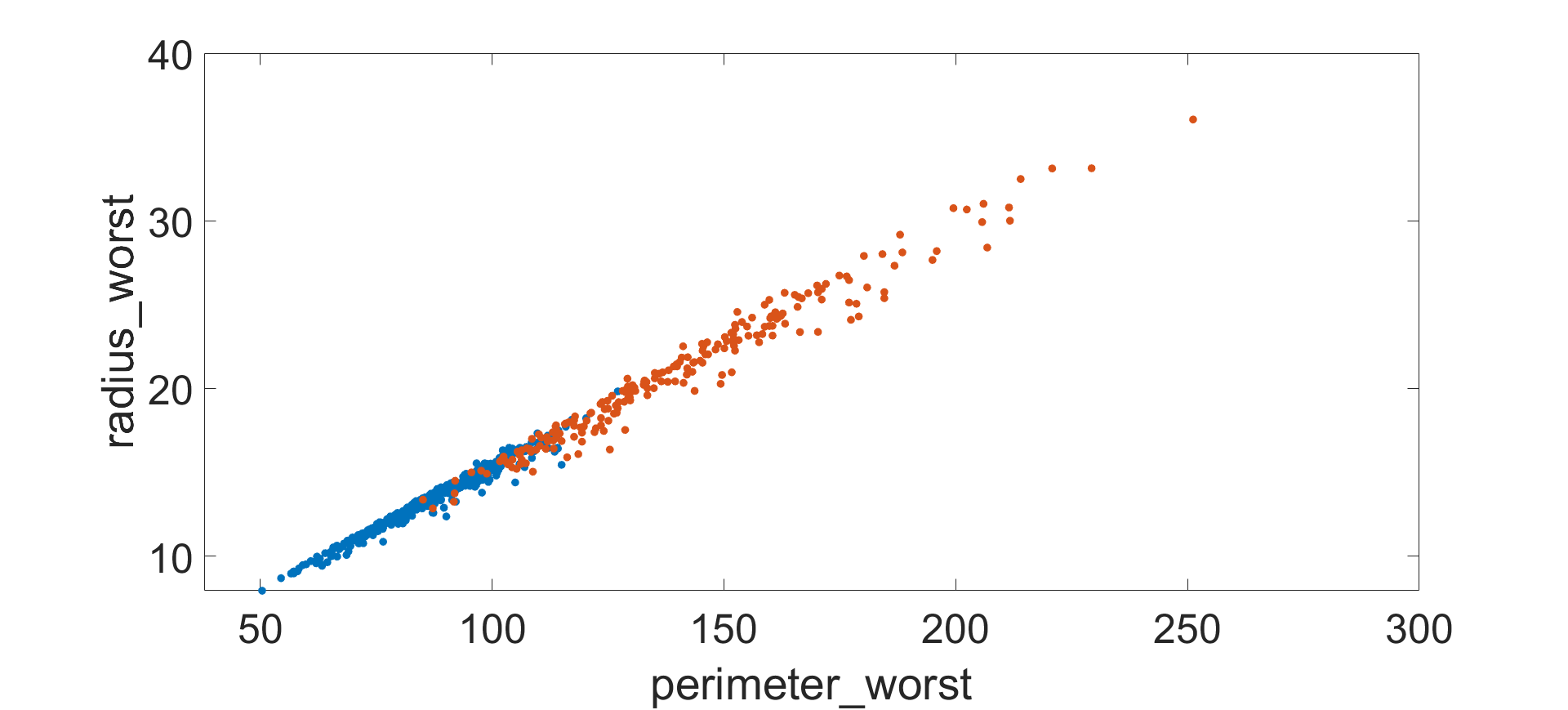}\label{fig:w2}}
\subfigure{\includegraphics[width=82mm]{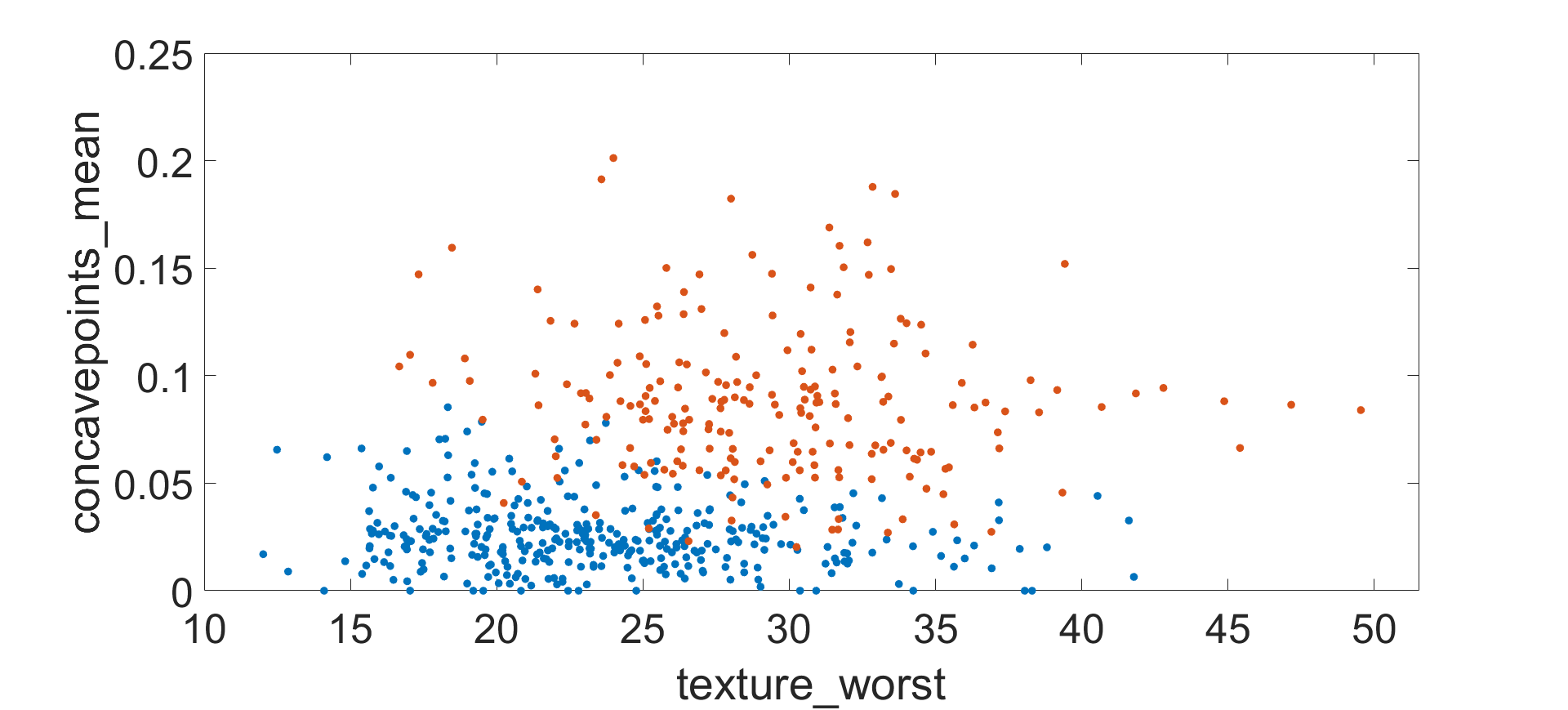}\label{fig:w3}}
\subfigure{\includegraphics[width=82mm]{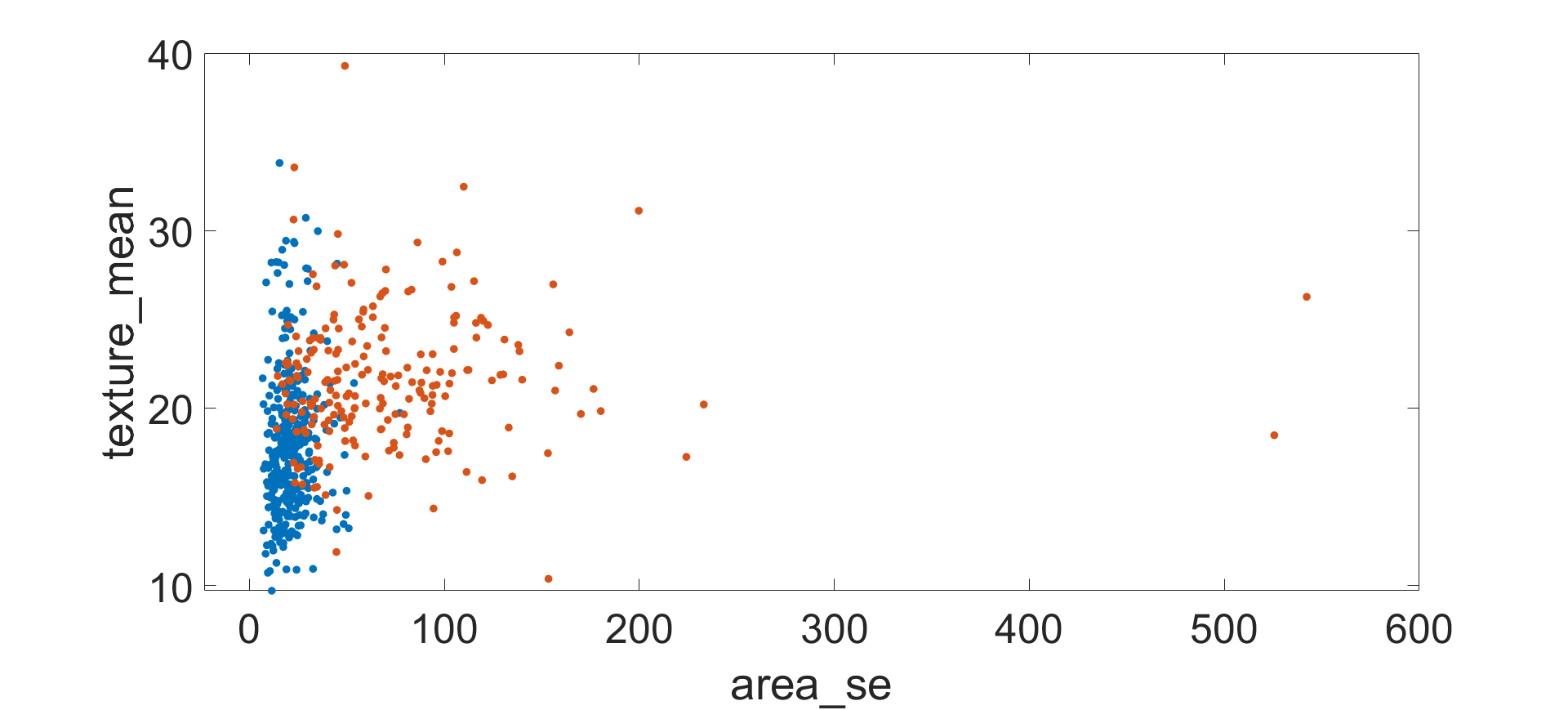}\label{fig:w4}}
\caption{t-SNE scatter plot examples of some initial features for malignant cancerous samples (red circles), and benign non-cancerous samples (blue circles). (a)~{Area worst vs. Concave points worst} y (b)~{Perimeter worst vs. Radius worst} (c)~{Texture worst vs. Concave point mean.} y (d)~{Area see vs. Texture mean.}}
\label{fig:images}
\end{figure}

For illustration, Fig.~\ref{fig:images} shows some feature examples from the dataset introduced in Section~\ref{ssec:data}. t-distributed Stochastic Neighbor Embedding (t-SNE) was used to yield the scatter plot between couples for malignant cancerous samples (red circles), and benign non-cancerous samples (blue circles). Note that samples can be separated linearly, allowing the use of linear-methods-based learning schemes.

\begin{figure}[htbp]
\centering
\subfigure[Cohen's d]{\includegraphics[width=60.5mm]{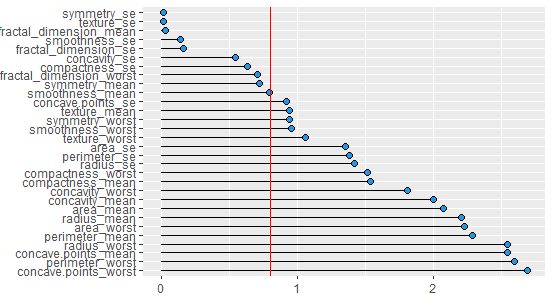}\label{fig:cohend}}
\subfigure[Cohen's D]{\includegraphics[width=60.5mm]{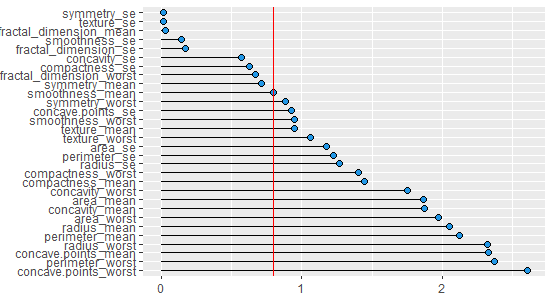}}\label{fig:cohendD}
\subfigure[Cohen's $U_1$]{\includegraphics[width=60.5mm]{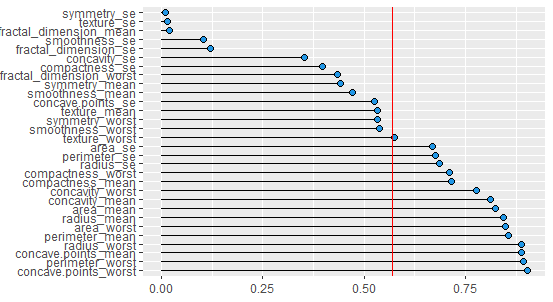}\label{fig:cohenu1}}
\subfigure[Cohen's $U_2$]{\includegraphics[width=60.5mm]{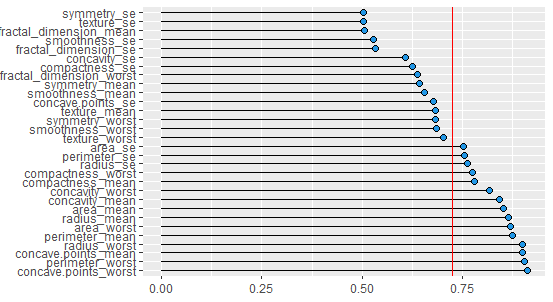}\label{fig:cohenu2}}
\subfigure[Cohen's $U_3$]{\includegraphics[width=60.5mm]{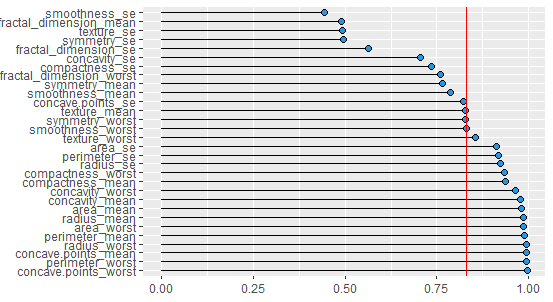}\label{fig:cohenU3}}
\caption{Decision rule threshold results are illustrated by a red line for the features under study. (a)~{Cohen's d: $0.8$.} y (b)~{Cohen's D: $0.8$.} (c)~{Cohen's $U_1$: $0.5$.} (d)~{Cohen's $U_2$: $0.7$.}(e)~{Cohen's $U_3$: $0.8$.}}
\label{fig:effects}
\end{figure}

Fig.~\ref{fig:effects} and Table~\ref{tab:effectsizes} show the effect size values observed for each feature using the different measures introduced in Section~\ref{ssec:effect} and the feature-selector-based decision rule, Section~\ref{ssec:rule}. Note that in Fig.~\ref{fig:effects} the Cohen's d and Cohen's D results are based on the decision rule when the effect size value is greater than $0.8$, while Cohen's $U_1$ to Cohen's $U_3$ results, the decision rule is based on the mean of its rank. A red vertical line in the figures shows the decision rule threshold. 
Decision rule results in Tables~\ref{tab:rank} and confidence intervals results in Table~\ref{tab:CI} complement the observed results of Fig.~\ref{fig:effects}. 

In Table~\ref{tab:effectsizes}, the Feature column contains the variables and the Effect Sizes columns show the more significant feature for each effect size measure based on the decision rule. The feature-selector is highlighted with a black $x$ and the common features for all effect sizes measure with a blue $x$. 

\begin{table}[htbp]
\caption{Effect size feature-selector-based learning results. In blue, common features for all effect size measures are reported.}
    \centering
    \scalebox{0.7}{
    \begin{tabular}{|c|c|c|c|c|c|c|}
    \hline
    &\multicolumn{5}{|c|}{\textbf{Effect Sizes}} \\
    \cline{2-6}
        \textbf{Features} &  \textbf{Cohen's d}&  \textbf{Cohen's D}&  \textbf{Cohen's $\bs{U_1}$}&  \textbf{Cohen's $\bs{U_2}$}& \textbf{Cohen's $\bs{U_3}$} & \textbf{Common features}\\
        \hline
         Radius mean&  $x$&  $x$&  $x$&  $x$& $x$ & $\hlb x$\\
         \hline
         Perimeter mean&  $x$&  $x$&  $x$& $x$ & $x$ & $\hlb x$\\
         \hline
         Area mean&  $x$&  $x$&  $x$&  $x$& $x$ & $\hlb x$\\
         \hline
         Te$x$ture mean&  $x$&  $x$&  &  & &\\
         \hline
         Compactness mean&  $x$&  $x$&  $x$&  $x$& $x$ & $\hlb x$\\
         \hline
         Smoothness mean&  &  &  &  & &\\
         \hline
         Concavity mean&  $x$&  $x$&  $x$&  $x$& $x$ & $\hlb x$\\
         \hline
         Concave points mean&  $x$&  $x$&  $x$&  $x$& $x$ & $\hlb x$\\
         \hline
         Symmetry mean&  &  &  &  & &\\
         \hline
         Fractal dimension mean&  &  &  &  & &\\
         \hline
         Radius se&  $x$&  $x$&  $x$&  $x$& $x$ &$\hlb x$\\
         \hline
         Perimeter se&  $x$&  $x$&  $x$&  $x$& $x$ & $\hlb x$\\
         \hline
         Area se&  $x$&  $x$&  $x$&  $x$& $x$ &$\hlb x$\\
         \hline
         Texture se&  &  &  &  & &\\
         \hline
         Compactness se&  &  &  &  & &\\
         \hline
         Smoothness se&  &  &  &  & &\\
         \hline
         Concavity se&  &  &  &  & &\\
         \hline
         Concave points se&  $x$&  $x$&  &  & $x$ &\\
         \hline
         Symmetry se&  &  &  &  & &\\
         \hline
         Fractal dimension se&  &  &  &  &  &\\
         \hline
         Radius worst&  $x$&  $x$&  $x$&  $x$& $x$ &$\hlb x$\\
         \hline
         Perimeter worst&  $x$&  $x$&  $x$&  $x$& $x$ &$\hlb x$\\
         \hline
         Area worst&  $x$&  $x$&  $x$&  $x$& $x$ &$\hlb x$\\
         \hline
         Texture worst&  $x$&  $x$&  $x$&  & $x$ &\\
         \hline
         Compactness worst&  $x$&  $x$&  $x$&  $x$& $x$ &$\hlb x$\\
         \hline
         Smoothness worst&  $x$&  $x$&  &  & $x$ &\\
         \hline
         Concavity worst&  $x$&  $x$&  $x$&  $x$& $x$ &$\hlb x$\\
         \hline
         Concave points worst&  $x$&  $x$&  $x$&  $x$& $x$ &$\hlb x$\\
         \hline
         Symmetry worst&  $x$&  $x$&  &  & &\\
         \hline
         Fractal dimension worst&  &  &  &  & &\\
         \hline
    \end{tabular}}
    \label{tab:effectsizes}
\end{table}

For feature-selector-based learning, a classical SVM with a linear kernel was used as a learner tool with 10-fold cross-validation. The experiment was repeated $20$ times randomly to assess consistency in the detection. The True Positive Rate or Recall or Sensitivity (TPR), False Positive Rate (FPS) or false alarm rate, Accuracy (ACC), and Area Under the Curve (AUC) classification performance metrics yield interesting results in detecting abnormalities in breast cancer. The metrics used in this study show excellent results, over $90\%$ for all the effect sizes studied, except for Cohen’s U1 with an accuracy of $61.18\%$, but an acceptable AUC. Also, all common features for all the effect sizes were tested yielding an excellent performance. Additionally, the feature-selector filter method Relief \cite{Kira1992} was tested for comparison with the effect size measures used, see Table~\ref{tab:metrics}. It is important to highlight that the complexity of effect sizes is linear $\mathcal{O}(f.n)$, while the Relief is quadratic $\mathcal{O}(f.n^2)$, where $f$ is the number of features and $n$ is the number of instances \cite{Urbanowicz2018}. These excellent results suggest that the effect size proposed is within the standards of the feature-selector methods. 

\begin{table}[htbp]
\caption{Ranks observed, Means, and Decision rules}
\begin{center}
\scalebox{0.9}{
\begin{tabular}{|c|c|c|c|c|}
\hline
\textbf{Effect Size} & \textbf{Mean} & \textbf{Rank} &\textbf{Decision Rule}\\  
\hline 
Cohen's d & $1.29$ & [$0, 2.7$] & $0.8$ \\
\hline
Cohen's D & $1.20$ & [$0, 2.6$] & $0.8$ \\
\hline
Cohen's $U_1$ & $0.56$ & [$0, 0.9$] & $0.5$\\
\hline
Cohen's $U_2$ & $0.74$ & [$0.5, 0.95$] & $0.7$\\
\hline
Cohen's $U_3$ & $0.84$ & [$0.3, 1$] & $0.8$\\
\hline 
\end{tabular}}
\label{tab:rank}
\end{center}
\end{table}

\begin{table}[htbp]
\caption{Confidence Intervals for the effect sizes studied.}
    \centering
    \scalebox{0.9}{
    \begin{tabular}{|c|c|c|c|c|c|}
    \hline
    &\multicolumn{5}{|c|}{\textbf{Confidence Interval}} \\
    \cline{2-6}
        \textbf{Features} &  \textbf{Cohen's d}&  \textbf{Cohen's D}&  \textbf{Cohen's $U_1$}&  \textbf{Cohen's $U_2$}& \textbf{Cohen's $U_3$}\\
        \hline
         Radius mean&  [$1.99, 2.41$]&  [$1.81, 2.31$]&  [$0.81, 0.87$]&  [$0.84, 0.88$]& [$0.97, 0.99$]\\
         \hline
         Texture mean&  [$0.76, 1.12$]&  [$0.77, 1.13$]&  [$0.45, 0.59$]& [$0.64, 0.71$]& [$0.77, 0.86$]\\
         \hline
         Perimeter mean&  [$2.07, 2.5$]&  [$1.88, 2.39$]&  [$0.82, 0.88$]&  [$0.85, 0.89$]& [$0.98, 0.99$]\\
         \hline
         Area mean&  [$1.86, 2.28$]&  [$1.62, 2.12$]&  [$0.78, 0.85$]&  [$0.82, 0.87$]& [$0.96, 0.98$]\\
         \hline
         Smoothness mean&  [$0.61, 0.96$]&  [$0.62, 0.97$]&  [$0.38, 0.64$]&  [$0.62, 0.68$]& [$0.73, 0.83$]\\
         \hline
         Compactness mean&  [$1.34, 1.72$]&  [$1.24, 1.66$]&  [$0.66, 0.75$]&  [$0.74, 0.8$]& [$0.91, 0.95$]\\
         \hline
         Concavity mean&  [$1.79, 2.2$]&  [$1.64, 2.12$]&  [$0.77, 0.84$]&  [$0.81, 0.86$]& [$0.96, 0.98$]\\
         \hline
         Concave points mean&  [$2.31, 2.76$]&  [$2.07, 2.61$]&  [$0.85, 0.9$]&  [$0.87, 0.91$]& [$0.98, 0.99$]\\
         \hline
         Symmetry mean&  [$0.54, 0.89$]&  [$0.53, 0.89$]&  [$0.35, 0.51$]&  [$0.6, 0.67$]& [$0.7, 0.81$]\\
         \hline
         Fractal dimension mean&  [$-0.14, 0.19$]&  [$-0.14, 0.19$]&  [$0, 0.14$]&  [$0.5, 0.53$]& [$0.42, 0.55$]\\
         \hline
         Radius se&  [$1.23, 1.61$]&  [$1.05, 1.49$]&  [$0.63, 0.73$]&  [$0.73, 0.78$]& [$0.89, 0.94$]\\
         \hline
         Texture se&  [$-0.15, 0.18$]&  [$-0.14, 0.18$]&  [$0, 0.13$]&  [$0.5, 0.53$]& [$0.42, 0.56$]\\
         \hline
         Perimeter se&  [$1.19, 1.56$]&  [$1.01, 1.45$]&  [$0.61, 0.72$]&  [$0.72, 0.78$]& [$0.88, 0.94$]\\
         \hline
         Area se&  [$1.16, 1.54$]&  [$0.96, 1.4$]&  [$0.61, 0.71$]&  [$0.71, 0.77$]& [$0.87, 0.93$]\\
         \hline
         Smoothness se&  [$-0.03, 0.3$]&  [$-0.02, 0.3$]&  [$0, 0.21$]&  [$0.5, 0.56$]& [$0.37, 0.51$]\\
         \hline
         Compactness se&  [$0.45, 0.8$]&  [$0.44, 0.8$]&  [$0.3, 0.47$]&  [$0.59, 0.65$]& [$0.67, 0.78$]\\
         \hline
         Concavity se&  [$0.36, 0.71$]&  [$0.4, 0.73$]&  [$0.25, 0.43$]&  [$0.57, 0.63$]& [$0.64, 0.76$]\\
         \hline
         Concave points se&  [$0.74, 1.1$]&  [$0.75, 1.11$]&  [$0.44, 0.58$]&  [$0.64, 0.7$]& [$0.77, 0.86$]\\
         \hline
         Symmetry se&  [$-0.15, 0.18$]&  [$-0.16, 0.19$]&  [$0, 0.13$]&  [$0.5, 0.53$]& [$0.42, 0.56$]\\
         \hline
         Fractal dimension se&  [$-0.008, 0.33$]&  [$-0.006, 0.33$]&  [$0, 0.23$]&  [$0.5, 0.56$]& [$0.49, 0.62$]\\
         \hline
         Radius worst&  [$2.31, 2.76$]&  [$2.07, 2.61$]&  [$0.85, 0.9$]&  [$0.87, 0.91$]& [$0.98, 0.99$]\\
         \hline
         Texture worst&  [$0.87, 1.24$]&  [$0.88, 1.25$]&  [$0.5, 0.63$]&  [$0.66, 0.73$]& [$0.81, 0.89$]\\
         \hline
         Perimeter worst&  [$2.37, 2.82$]&  [$2.11, 2.66$]&  [$0.86, 0.91$]& [$0.88, 0.92$]& [$0.991, 0.997$]\\
         \hline
         Area worst&  [$2.01, 2.44$]&  [$1.72, 2.24$]&  [$0.81, 0.87$]&  [$0.84, 0.88$]& [$0.97, 0.99$]\\
         \hline
         Smoothness worst&  [$0.78, 1.13$[&  [$0.76, 1.13$]&  [$0.46, 0.6$]&  [$0.65, 0.71$]& [$0.78, 0.87$]\\
         \hline
         Compactness worst&  [$1.32, 1.7$]&  [$1.19, 1.62$]&  [$0.65, 0.75$]&  [$0.74, 0.8$]& [$0.9, 0.95$]\\
         \hline
         Concavity worst&  [$1.61, 2.01$]&  [$1.54, 1.98$]&  [$0.73, 0.81$]&  [$0.78, 0.84$]& [$0.94, 0.97$]\\
         \hline
         Concave points worst&  [$2.46, 2.92$]&  [$2.35, 2.87$]&  [$0.87, 0.92$]&  [$0.89, 0.92$]& [$0.993, 0.998$]\\
         \hline
         Symmetry worst&  [$0.76, 1.12$]&  [$0.69, 1.08$]&  [$0.45, 0.59$]&  [$0.64, 0.71$]& [$0.77, 0.86$]\\
         \hline
         Fractal dimension worst&  [$0.53, 0.88$]&  [$0.48, 0.85$]&  [$0.34, 0.5$]&  [$0.6, 0.67$]& [$0.7, 0.81$]\\
         \hline
    \end{tabular}}
    \label{tab:CI}
\end{table}

\begin{table}[htbp]
\caption{True Positive Rate or Recall (TPR), False Positive Rate (FPS), Accuracy (ACC), and Area Under the Curve (AUC) classification performance metrics with this dataset}
\begin{center}
\begin{tabular}{|c|c|c|c|c|c|c|}
\hline
\textbf{\textit{Features}} & \textbf{\textit{TPR}}& \textbf{\textit{FPS}} & \textbf{\textit{ACC}} & \textbf{\textit{AUC}}\\
\hline
 Cohen's d & 97.20& 92.06& 95.29 & 0.98\\
\hline
 Cohen's D & 90.65& 90.48& 90.59 & 0.98\\
\hline
 Cohen's $U_1$ & 72.90& 41.27& 61.18 & 0.88\\
\hline
 Cohen's $U_2$ & 98.13& 84.13& 92.94 & 0.98\\
\hline
 Cohen's $U_3$ & 96.47& 99.07& 92.06 & 0.98\\
\hline
Common features & 94.39 &93.65 &94.12 & 0.97\\
\hline
Relief & 100 & 95.24& 98.24 & 0.99\\
\hline
\end{tabular}
\label{tab:metrics}
\end{center}
\end{table}

\section{Conclusions}
\label{sec:con}
In this work, a statistical feature-selector-based learning tool based on effect sizes was proposed to detect abnormalities in breast cancer from features extracted from cell nuclei images. Five parametric methods based on Cohen’s d, Cohen’s D, Cohen’s $U_1$, Cohen’s $U_2$, and Cohen’s $U_3$ were used as feature-selectors to estimate a dimensional reduction of the data based on a numeric decision rule. To assess the potentiality of the tool proposed a classical SVM with a linear kernel was used as a learner classifier with a 10-fold cross-validation. The experiment was repeated 20 times randomly to assess consistency in detecting abnormalities in breast cancer. The True Positive Rate or Recall or Sensitivity (TPR), False Positive Rate (FPS) or false alarm rate, Accuracy (ACC), and Area Under the Curve (AUC) were used as metric performance achieving excellent results, over $90\%$ for all the effect sizes measures studied, except for Cohen’s U1 with accuracy over $60\%$, but an acceptable AUC. These excellent results suggest that the effect size of statistical feature-selector-based learning to detect breast cancer is within the standards of the feature-selector methods.
A notable advantage of using effect size as feature-selection-based learning is the lower computational complexity and versatility. The effect size measures could be treated as a filter method, so the complexity is lower than other methods when the data dimension is high because they are estimated directly from the data and are independent of the sample size. The main limitation lies in the statistical significance. It can be misleading if it is influenced by sample size, since increasing the sample size may increase the probability of finding a statistically significant effect.
Future work will focus on a comprehensive evaluation of the proposed approach with parametric and non-parametric effect size measures as feature-selection-based learning and on deriving instances of the method with other datasets tailored for specific medical applications in detecting abnormalities in breast cancer.


\end{document}